# Analyzing Uniaxial Compressive Strength of Concrete Using a Novel Satin Bowerbird Optimizer

Hossein Moayedi, Amir Mosavi
School of the Built Environment, Oxford Brookes University, Oxford, UK
a.mosavi@brookes.ac.uk

**Abstract**
Surmounting the complexities in analyzing the mechanical parameters of concrete entails selecting an appropriate methodology. This study integrates artificial neural network (ANN) with a novel metaheuristic technique, namely satin bowerbird optimizer (SBO) for predicting uniaxial compressive strength (UCS) of concrete. For this purpose, the created hybrid is trained and tested using a relatively large dataset collected from the published literature. Three other new algorithms, namely Henry gas solubility optimization (HGSO), sunflower optimization (SFO), and vortex search algorithm (VSA) are also used as benchmarks. After attaining a proper population size for all algorithms, the Utilizing various accuracy indicators, it was shown that the proposed ANN-SBO not only can excellently analyze the UCS behavior, but also outperforms all three benchmark hybrids (i.e., ANN-HGSO, ANN-SFO, and ANN-VSA). In the prediction phase, the correlation indices of 0.87394, 0.87936, 0.95329, and 0.95663, as well as mean absolute percentage errors of 15.9719, 15.3845, 9.4970, and 8.0629%, calculated for the ANN-HGSO, ANN-SFO, ANN-VSA, and ANN-SBO, respectively, manifested the best prediction performance for the proposed model. Also, the ANN-VSA achieved reliable results as well. In short, the ANN-SBO can be used by engineers as an efficient non-destructive method for predicting the UCS of concrete.

## 1 Introduction

Determining the uniaxial compressive strength (UCS) is a straightforward method for evaluating the strength of concrete. Since direct methods are costly [1] and incur damages to the structure [2], many scholars have turned to indirect intelligent techniques like artificial neural network (ANN) [3] and support vector machine (SVM) [4].

In the literature, plenty of research items have been dedicated to the intelligent simulation of concrete characteristics. Han et al. [5] presented a promising performance of random forest (RF) for predicting the CS of high-performance concrete. The best relative error was around 11.7%. Nguyen et al. [6]

tested the feasibility of conventional and second-order ANN for estimating the CS of foamed concrete. They finally professed the suitability of these methods for optimal mixture design. Tinoco et al. [7] proved the competency of SVM in analyzing the UCS of jet grouting columns. Zhang et al. [8] employed an RF regression for appraising the UCS of lightweight concrete. With around 97% correlation, the proposed model was found to be accurate for this purpose. The same method was developed for the shear strength [9]. Ozcan et al. [10] investigated the effect of waste tire rubber powder and blast furnace slag on the CS of cement mortars using Ada boost and SVM models. Regarding the suitable accuracy indices, i.e., the coefficient of determination ($R^2$) and MAPE of 0.9831 and 0.1105, the Ada boost was introduced as a capable approximator. More studies concerning machine learning applications can be found in Ref.s [11-14].

By doing comparative studies, scholars have sought the most potent predictive models. Dutta et al. [15] conducted a comparison among well-known data mining techniques, namely multi adaptive regression spline (MARS), Gaussian process regression (GPR), and minimax probability machine regression (MPMR) applied to CS modeling. While all three predictors attained a high accuracy of prediction, based on the correlation values (0.9485, 0.9570, and 0.9352 for the GPR, MARS, and MPMR, respectively), the performance of the MARS was more promising. In a similar effort, Jalal et al. [16] compared the robustness of five popular intelligent approaches, namely ANN, nonlinear multivariable regression (NMVR), genetic programming (GP), adaptive neuro-fuzzy inference system (ANFIS), and SVM for estimating the CS of optimized recycled rubber concrete. With a root mean square error (RMSE) of 1.393 and the $R^2$ of 0.989, the SVM surpassed other methods. However, the formula of the ANN and GP were applicable. The extreme gradient boosting suggested by Nguyen-Sy et al. [17] could predict the CS of concrete with better performance than the ANN and SVM. In the mentioned work, the mixture ingredients and the age were considered as influential factors.

Metaheuristic optimizers have provided effective solutions to many engineering analyses suchlike geotechnical issues [18], environmental risks [19], energy efficiency [20], etc. Concerning the simulation of concrete parameters, scholars have used famous algorithms like particle swarm optimization (PSO) for shear strength [21], firefly algorithm (FA) for creep strain [22], ant lion optimization (ALO) for slump [23] modeling. Zhang and Wang [24] proposed an optimized version of the least square support vector regression (LSSVR) for analyzing the bond strength of composite joints (fiber-reinforced polymers). The optimizer was beetle antennae search algorithm synthesized with the Levy flight technique. Due to the obtained satisfying results (Correlation = 0.983 and RMSE = 1.99), the proposed model was introduced as a capable approach. The same metaheuristic algorithm was applied to the SVR by Sun et al. [25] for approximating the UCS and the permeability coefficient of concrete. With a close-to-ideal correlation (> 0.99) obtained for both learning and prediction phases, the proposed model performed successfully in their research. In research by Bui et al. [26], the ANN was successfully tuned by a nature-inspired optimizer called whale optimization algorithm (WOA). With 0.8976 correlation, this algorithm exceeded dragonfly (0.8209) and ant colony algorithm (0.8000). Shamshirband et al. [27] combined the SVR with bat algorithm for simulating the unconfined compressive strength of bricks made of cement. Since this hybrid model performed better than conventional ANN and ANFIS, it was denoted as a capable approximator. Ma et al. [28] addressed the optimization of ANN using salp swarm algorithm (SSA) for the compressive strength prediction. The created hybrid was better than those developed by algorithms like artificial bee colony and shuffled frog leaping algorithm. Huang et al. [29] suggested using a firefly-tuned SVR for modeling the UCS and flexural strength of steel fibre reinforced concrete. Regarding the calculated correlations (85 and 91% for the flexural and compressive strengths, respectively), they concluded that firefly is a promising option for adjusting the SVR.

The excellent efficiency of metaheuristic approaches in handling complex computations encouraged the authors to employ a novel technique, namely satin bowerbird optimizer (SBO) for optimizing an ANN used to predict the UCS. Moreover, three newly-developed search schemes of Henry gas solubility optimization (HGSO), sunflower optimization (SFO), and vortex search algorithm (VSA) are also assigned to the same task as benchmark optimizers. Such techniques train the ANN using advanced search strategies that can provide optimal solutions to any intricate problem.

## 2 Data and statistical analysis

Zhao et al. [30] collected a comprehensive concrete dataset from the published literature. This work uses the records of the USC observed for different values of compressive strength of cement (CSC), tensile strength of cement (TSC), curing age (CA), the maximum size of the crushed stone ($D_{max}$), stone powder content (SPC), fine modulus (FM), the ratio of water to binder (W/B), and sand ratio (SR). In other words, the CSC, TSC, CA, $D_{max}$, SPC, FM, W/B, and SR play the role of input parameters for the USC [31]. These factors were available for 323 samples. Figure 1 shows the distribution of the USC versus the influencing factors.

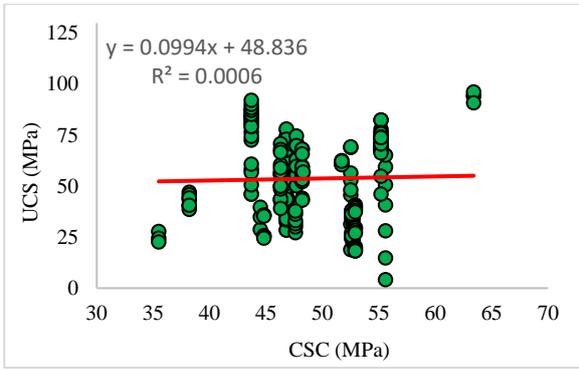
(a)

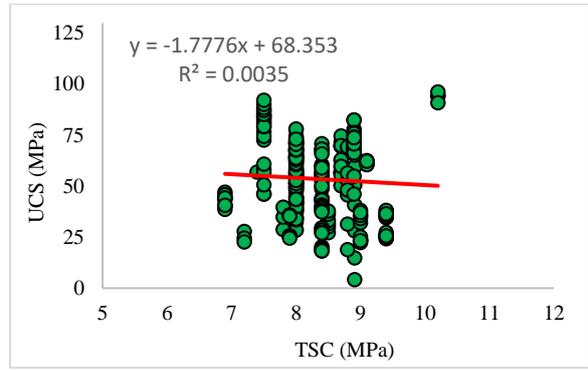
(b)

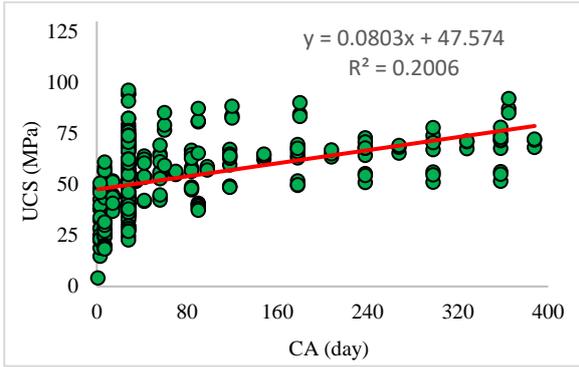
(c)

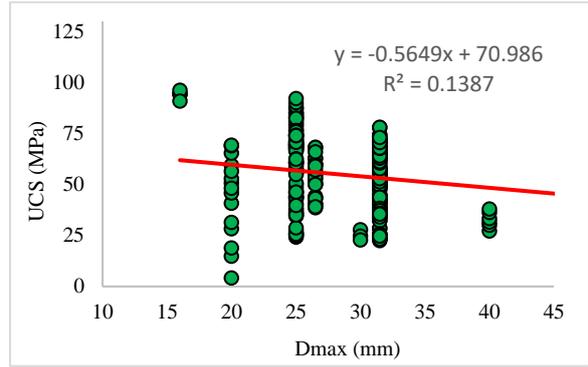
(d)

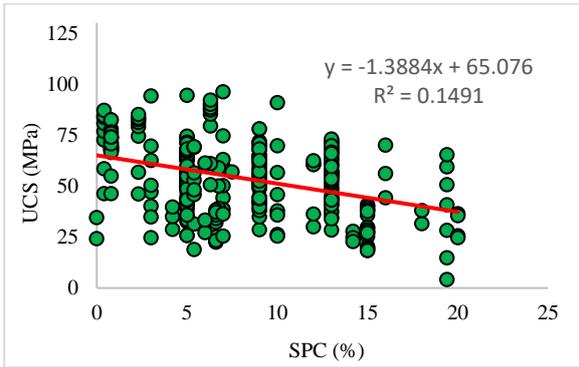
(e)

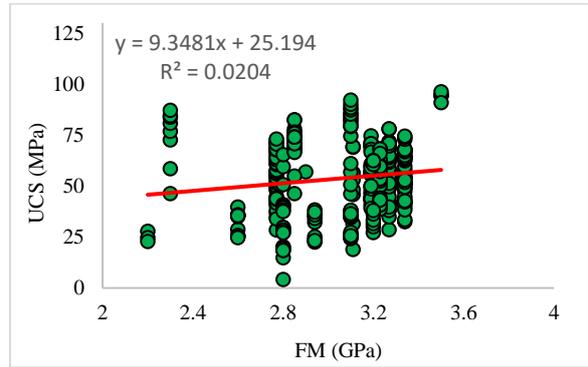
(f)

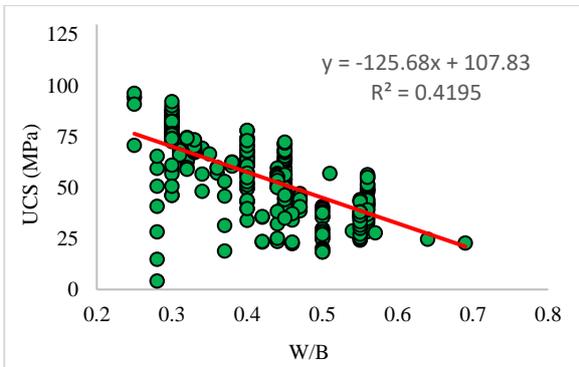
(g)

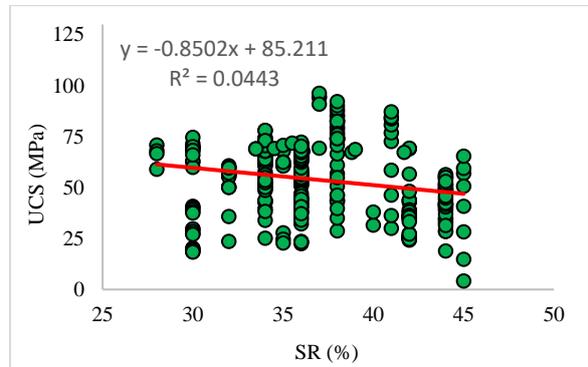
(h)

Figure 1: The distribution of UCS versus the inputs.

Moreover, statistical indicators presented in Table 1 describe the used data. As is seen, the UCS varies in a relatively wide range (i.e., from 4.23 to 96.30 $MPa$). Also, the age of the samples receives twenty-six different values varying from 1 to 388 days where the largest frequency belongs to 28-day samples.

Table 1: Statistical indicators of the UCS and input parameters.

| Parameter | Unit | Descriptive indicator | | | | |
|---|---|---|---|---|---|---|
| | | Mean | Standard Error | Sample Variance | Minimum | Maximum |
| CSC | MPa | 48.35 | 0.24 | 18.79 | 35.50 | 63.40 |
| TSC | MPa | 8.28 | 0.03 | 0.34 | 6.90 | 10.20 |
| CA | Day | 75.58 | 5.47 | 9667.60 | 1.00 | 388.00 |
| D$_{max}$ | mm | 30.71 | 0.65 | 134.97 | 16.00 | 80.00 |
| SPC | % | 8.24 | 0.27 | 24.01 | 0.00 | 20.00 |
| FM | GPa | 3.04 | 0.01 | 0.07 | 2.20 | 3.50 |
| W/B | - | 0.43 | 0.01 | 0.01 | 0.25 | 0.69 |
| SR | % | 37.13 | 0.24 | 19.04 | 28.00 | 45.00 |
| UCS | MPa | 53.64 | 0.98 | 310.48 | 4.23 | 96.30 |

## 3 Methodology

### 3.1 The SBO technique

Moosavi and Bardsiri [32] designed the SBO based on the life of bowerbirds. In the autumn and winter, these birds move to open forests to find food. The male birds attract a mate by building bowers and also stealing/destroying the bower of their neighbors. The female birds visit different bowers to select their mates.

In the SBO, a population is firstly created in a random way. It represents a collection of positions for bowers each of which contains a D-dimensional vector. The algorithm aims to optimize the parameters in these vectors. The fitness ($Fit_i$) of each bower is calculated as follows:

$$Fit_i = \begin{cases} \frac{1}{1+f(X_i)} & f(X_i) \geq 0 \\ 1+|f(X_i)| & f(X_i) < 0 \end{cases} \tag{1}$$

where $f(X_i)$ represents the cost of the bower.

Next, Given NB as the number of bowers, Equation 2 is used to calculate their probability.

$$P_i = \frac{Fit_i}{\sum_{n=1}^{N_B} Fit_i} \tag{2}$$

Based on this relationship, the bower with the highest fitness is selected as the best one. The rest of the population tries to mimic the best one to update themselves accordingly. Equations 3 and 4 are used for this purpose.

$$X_{ik}^{new} = X_{ik}^{old} + \lambda_k \left[\left(\frac{X_{jk} + X_{best,k}}{2}\right) - X_{ik}^{old}\right] \quad (3)$$

$$\lambda_k = \frac{a}{1 + P_j} \quad (4)$$

where $\lambda_k$ addresses the magnitude of step and $a$ stands for the greatest step size. Also, $j$ is given by the roulette wheel strategy. The next step is to apply the mutation. To this end, $X_{ik}$ experiences some random changes with a given probability. Based on Equations 5 and 6, a normal distribution (N) is exerted.

$$X_{ik}^{new} \sim N(X_{ik}^{old}, \sigma^2) = X_{ik}^{old} + (\sigma N(0,1)) \quad (5)$$

$$\sigma = z(Var_{max} - Var_{min}) \quad (6)$$

in which, $\sigma^2$ is the variance factor. Also, $Var_{min}$ and $Var_{max}$ denote the lower and upper bounds, respectively, and $z$ gives the relative (%) difference between these two parameters. Lastly, the algorithm selects the best bower as the solution after sorting new and old bowers [33]. More details about the SBO can be found in Ref.s [34; 35].

## 3.2 Benchmark techniques

Hashim et al. [36] introduced the HGSO based on physical behavior represented by Henry's law [37]. According to this law, the temperature is an important factor when gas is dissolved in a liquid in certain conditions. Moreover, the partial pressure of the gas can directly influence the solubility. In the HGSO algorithm, after initializing the population within the bounds, each individual (i.e., gas) is distinguished by its position and partial pressure value. In the following, based on the type of gases, they are classified into several groups and the elite gas is selected as the one with the largest equilibrium state. As the algorithm goes on, it adjusts the solubility and position of the gases [36].

In the SFO, the individuals are a number of artificial sunflowers. This kind of flower orients toward the sun in the morning and the opposite orientation in the evening. Gomes et al. [38] designed the SFO algorithm based on the inverse square law of radiation. According to this law, the severity of radiation is adversely proportional to the squared value of the distance between the sun and flower. Similar to other optimization techniques, the individuals mimic the elite one to improve their solutions. Note that, the elite flower is the one with the most potential orientation [39].

The third benchmark algorithm is the VSA. Doğan and Ölmez [40] designed this algorithm as a simulation of natural vortices structure. When the population is generated by using a Gaussian distribution, several circles search the surrounding space with respect to the initial center point and a primary radius. The local and global optimum responses are separately sought in the VSA. In this sense, an explorative behavior is taken in the initial steps to increase the potential of global search. By the time a promising solution is discovered, it is tuned by an exploitative operation. The solution is iteratively compared to the earlier one. In other words, the current solution is kept unless a solution with a lower error is discovered [41].

The banchmark algorithms are better explained in the earlier literature (HGSO [42; 43], SFO [44; 45], and VSA [41; 46]).

## 4 Results and discussion

For implementing the models, the first step is to designate the training and testing data. It is famously known that in any intelligent model, training algorithms go through the corresponding dataset to understand how the target is affected by a set of inputs. In this work, 80% of data (i.e., 258 out of 323 samples) is randomly chosen for this purpose and the remainder 20% (i.e., 65 out of 323 samples) is used to assess how generalizable the created pattern is. This group is called tested data.

For accuracy assessment, four criteria, namely the mean absolute error (MAE), mean absolute percentage error (MAPE), RMSE, and Pearson correlation coefficient (R) are defined based on the below equations.

$$MAE = \frac{1}{Z}\sum_{i=1}^{Z} |UCS_{i_{expected}} - UCS_{i_{predicted}}| \tag{7}$$

$$MAPE = \frac{1}{Z}\sum_{i=1}^{Z} |\frac{UCS_{i_{expected}} - UCS_{i_{predicted}}}{UCS_{i_{expected}}}| \times 100 \tag{8}$$

$$RMSE = \sqrt{\frac{1}{Z}\sum_{i=1}^{Z} [(UCS_{i_{expected}} - UCS_{i_{predicted}})]^2} \tag{9}$$

$$R = \frac{\sum_{i=1}^{Z}(UCS_{i_{predicted}} - \overline{UCS}_{predicted})(UCS_{i_{expected}} - \overline{UCS}_{expected})}{\sqrt{\sum_{i=1}^{Z}(UCS_{i_{predicted}} - \overline{UCS}_{predicted})^2}\sqrt{\sum_{i=1}^{Z}(UCS_{i_{expected}} - \overline{UCS}_{expected})^2}} \tag{10}$$

where $Z$ denotes the number of concrete samples in the intended dataset, and $UCS_{i_{predicted}}$ and $UCS_{i_{expected}}$ represent the simulated and expected UCSs, respectively.

### 4.1 Neural training and model development

The first step of applying metaheuristic techniques to a predictive model is defining the problem function. In the case of ANN, this function represents the general format of neural calculations with tunable weights and biases. A cost function is also defined which is the RMSE of the training process in this work. The algorithm tries to tune the neural interactions toward minimizing the cost function over a certain number of iterations. The decrease in the cost function is typically reflected by a convergence curve.

The effect of population size ($S_P$) on the performance of metaheuristic algorithms has been stated by many researchers [47-49]. Therefore, a trial and error practice on an appropriate range of $S_P$ (i.e., the $S_{PS}$ of 10, 50, 100, 200, 300, 400, and 500) yielded the best number of search agents for the given problem. The HGSO, SFO, VSA, and, SBO were tested with these values; and it was shown that the VSA with $S_P = 400$ trains the ANN with the highest accuracy, while this value was 300 for three other algorithms. Figure 2 shows the convergence curves of the selected ANN-HGSO, ANN-SFO, ANN-VSA, and ANN-SBO. As is seen, the algorithms had one thousand iterations to minimize the learning error. This figure also says that there is a considerable gap between the curves of both VSA and SBO with those of the HGSO and SFO. The lowest error, however, is achieved by the SBO-based model.

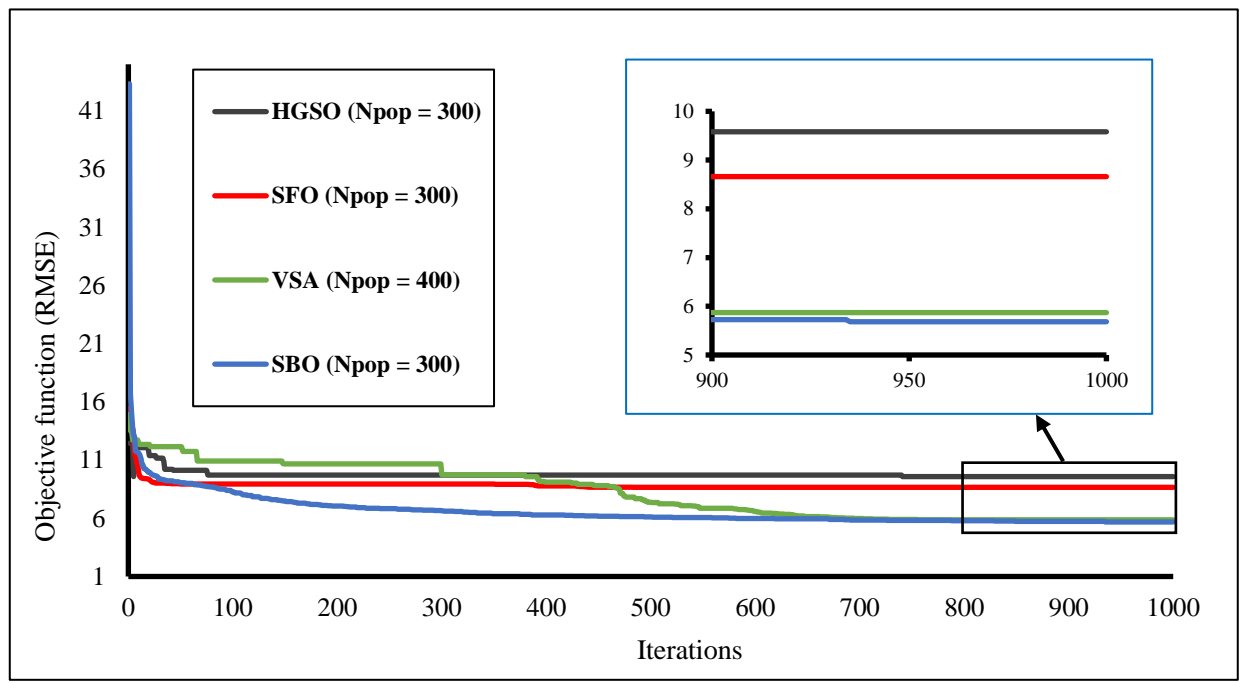

Figure 2: The convergence proceeding of the used optimizers.

The HGSO, SFO, VSA, and, SBO could train the ANN with the RMSEs of 9.5808, 8.6609, 5.8703, and 5.6826, respectively. the corresponding MAEs (MAPEs) were 7.5026 (19.8959%), 6.5996 (18.2992%), 4.4869 (12.4676%), and 4.1476 (11.8997%). Figure 3 shows the correlation between the training results.

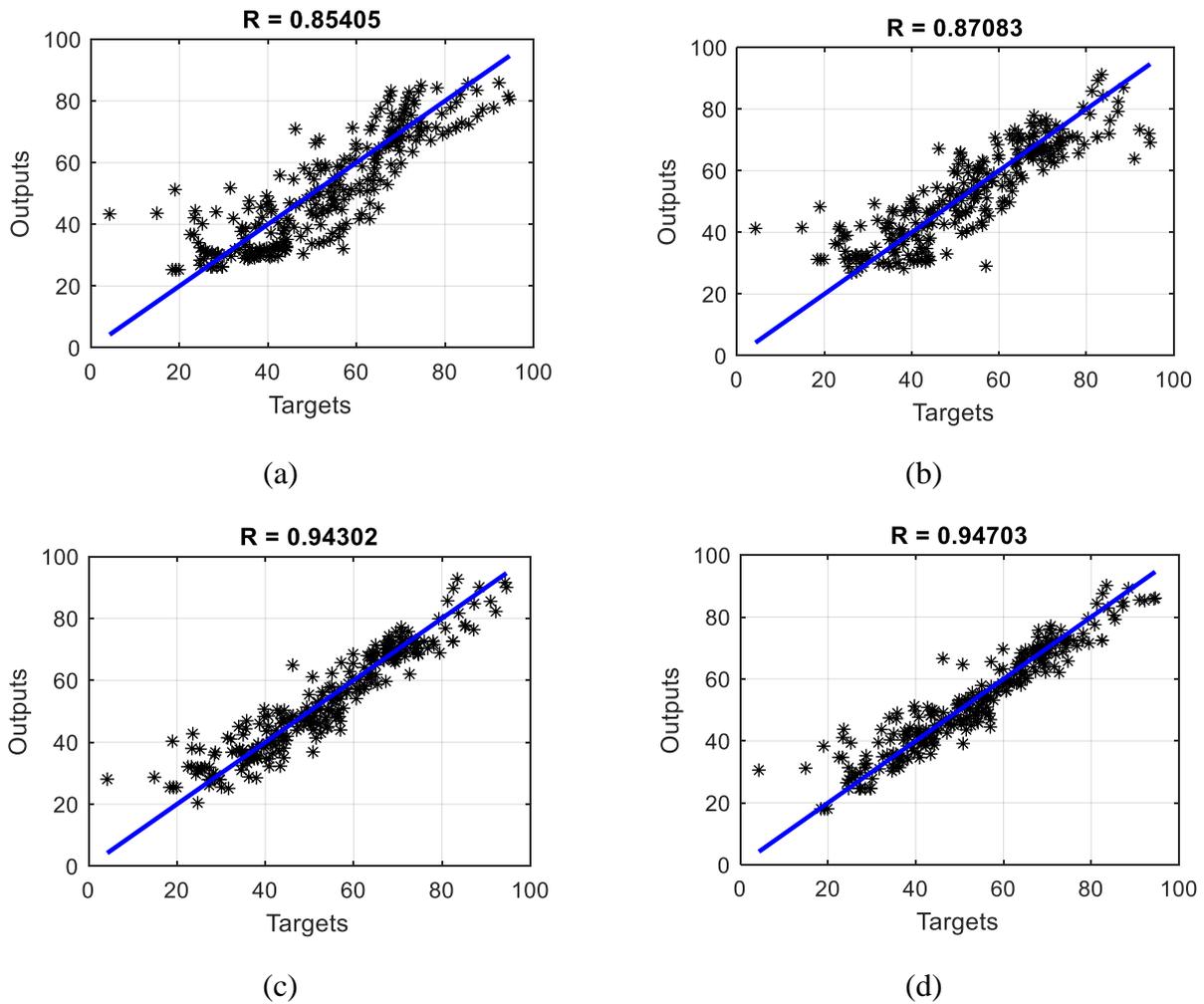

Figure 3: The correlation of the training results by (a) ANN-HGSO, (b) ANN-SFO, (c) ANN-VSA, and (d) ANN-SBO.

Having a glance at the presented regression charts, the results of all four ensembles are in a good agreement with the target UCS values. It can be also numerically demonstrated by the R values of 0.85405, 0.87083, 0.94302, and 0.94703. Famously, the goodness of the training results demonstrates the capability of the used trainers. Accordingly, it can be claimed that all four algorithms provided suitable training for the ANN.

The above results manifest the potential of search strategies embedded in the HGSO, SFO, VSA, and SBO. Without any supervision, these algorithms could test a huge number of solutions and move toward a better one regularly. It is worth noting that the initial response of the algorithms comprises a random solution. It is why the value of the objective function is very large in the beginning.

## 4.2 Prediction of the UCS

In the prediction phase, the knowledge of the models, that is acquired by analyzing training data, is tested by the second group of data (i.e., 65 testing samples). In this sense, the inputs are given to the models and the UCSs are predicted. They are then compared to the target values to assess the accuracy of prediction.

Figure 4 shows the results of the prediction. In the first parts, it compares the patterns of the target and output UCS, and along with that, the magnitude of the error (= $UCS_{i_{expected}} - UCS_{i_{predicted}}$) is graphically shown for all samples. It is well seen that the general behavior of the UCS has been nicely predicted by the models. Correctly following ups and downs indicates that the models have understood how the UCS changes by changes in input factors.

The values of accuracy indices are also given in Figure 4. The RMSEs were 9.5249, 8.5728, 5.3086, and 5.1679 which indicate a reliable prediction. Moreover, the calculated MAEs were 7.8632, 7.0550, 4.4006, and 3.9068 that along with their relative values (i.e., the MAPEs of 15.9719, 15.3845, 9.4970, and 8.0629%) support the goodness of the testing results. In terms of correlation, the models achieved the R values of 0.87394, 0.87936, 0.95329, and 0.95663.

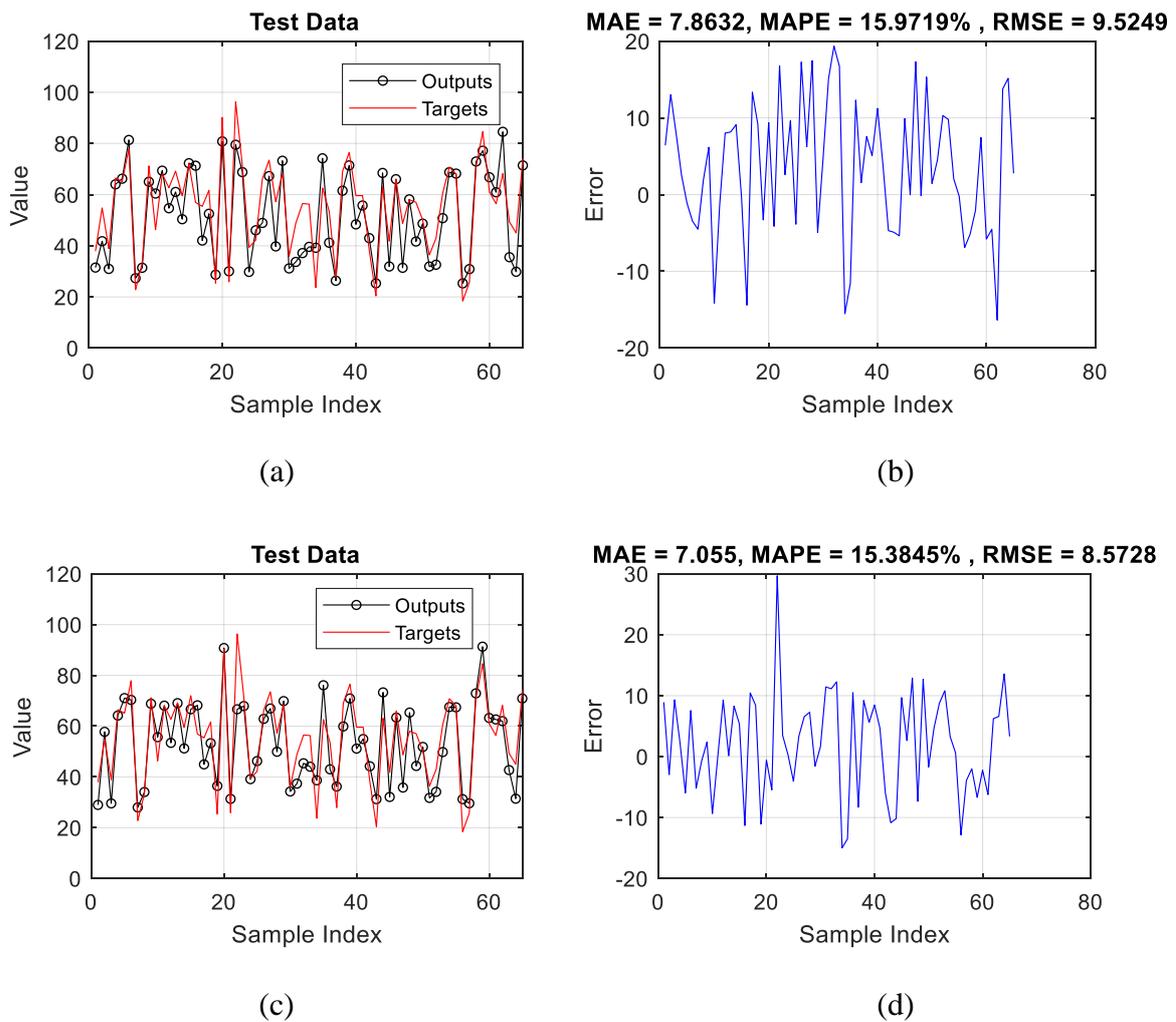

(a)　　　　　　　　　　　　　　(b)

(c)　　　　　　　　　　　　　　(d)

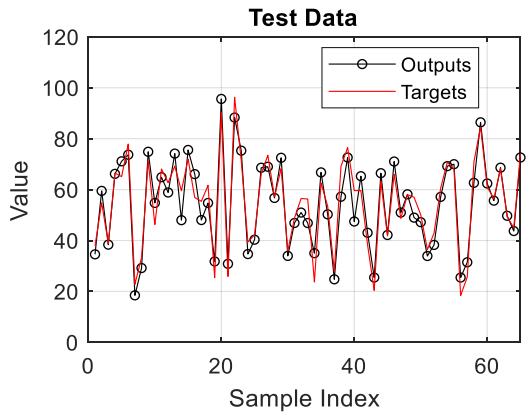 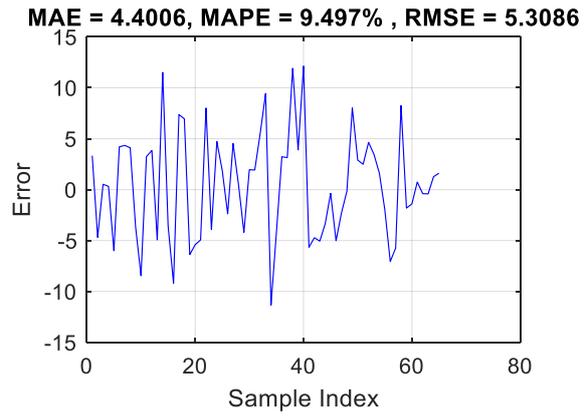

(e) (f)

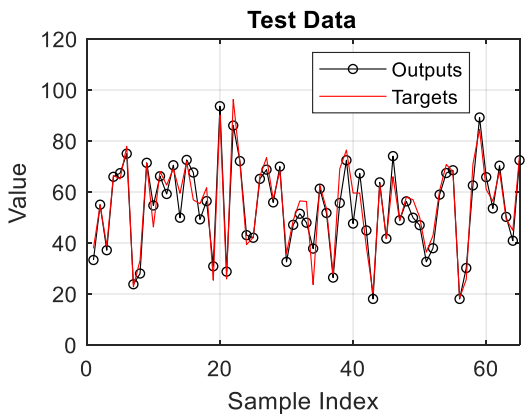 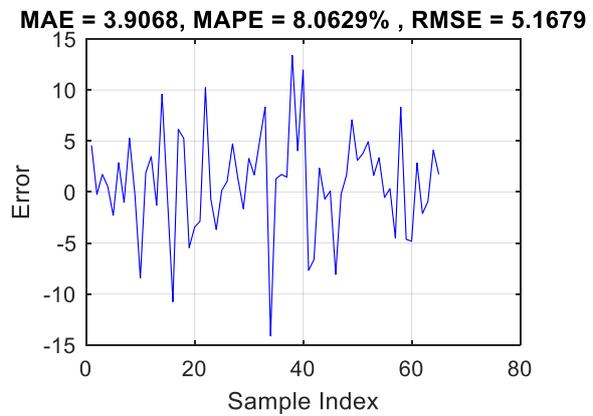

(g) (h)

Figure 4: Testing results in terms of (a, c, e, and g) pattern comparison and (b, d, f, and h) magnitude of the corresponding errors for the prediction of ANN-HGSO, ANN-SFO, ANN-VSA, and ANN-SBO, respectively.

### 4.3 Comparison

Referring to the results, the prediction of UCS was successfully done by the proposed models. However, the examined accuracies revealed that there are appreciable differences between the performance of the models in both phases. Hence, a comparison section is considered in this study to introduce the most capable hybrid predictor.

In the training phase, the potential of the VSA and SBO were quite superior over the HGSO and SFO. Apart from the accuracy indicators, it was also demonstrated by lower objective functions in the convergence curves. From this, it can be professed that the neural interactions optimized by the VSA and SBO are of higher quality. Consequently, the products of the ANN-VSA and ANN-SBO were distinguished by $R > 0.94$.

Not surprisingly, these methods were more accurate in the testing phase, too. For example, Figure 4 illustrates that the patterns predicted by the ANN-VSA and ANN-SBO are more sensitive to changes in the real pattern, compared to those of ANN-HGSO and ANN-SFO. Moreover, considering the relative errors, the MAPEs of the first two algorithms were below 10%, this index reported above 15% error for the ANN-HGSO and ANN-SFO.

To prepare a ranking of the models, they are compared in terms of all four indices (i.e., RMSE, MAE, MAPE, and R). Table 2 gives the obtained values altogether. As is seen, without any exception, the SBO attains the best accuracy (i.e., the smallest error indices and the largest correlation) in both phases, followed by the VSA, SFO, and HGSO.

Table 2: The summary of the used accuracy indicators.

| Hybrid | Results | | | | | | | |
|---|---|---|---|---|---|---|---|---|
| | Training phase | | | | Testing phase | | | |
| | RMSE | MAPE | MAE | R | RMSE | MAPE | MAE | R |
| ANN-HGSO | 9.5808 | 19.8959 | 7.5026 | 0.85405 | 9.5249 | 15.9719 | 7.8632 | 0.87394 |
| ANN-SFO | 8.6609 | 18.2992 | 6.5996 | 0.87083 | 8.5728 | 15.3845 | 7.0550 | 0.87936 |
| ANN-VSA | 5.8703 | 12.4676 | 4.4869 | 0.94302 | 5.3086 | 9.4970 | 4.4006 | 0.95329 |
| ANN-SBO | 5.6826 | 11.8997 | 4.1476 | 0.94703 | 5.1679 | 8.0629 | 3.9068 | 0.95663 |

## 5    ANN-SBO formula

The most reliable approximation of the UCS was achieved by the ANN trained by the SBO metaheuristic algorithm. As is known, neural interactions are established all over the network by a series of biases and weights that, altogether, aim to relate the UCS to the CSC, TSC, CA, $D_{max}$, SPC, FM, W/B, and SR. The formula given in Equation 11 expresses how the UCS is predicted by the ANN-SBO.

$$UCS = [LW] \cdot \left(\frac{2}{1 + e^{-2(([IW] \cdot [Input]) + [b1])}} - 1\right) + [b2] \tag{11}$$

where

$$LW = [0.2191 \quad 0.7574 \quad -0.1970 \quad -0.6422] \tag{12}$$

$$IW = \begin{bmatrix} 0.6833 & -0.5114 & -0.5029 & 0.2177 & 0.6053 & -0.8074 & 0.5226 & -0.6722 \\ 0.3366 & -0.7136 & 0.8009 & 0.4262 & -0.8098 & -0.7488 & -0.2934 & 0.1542 \\ 0.5897 & -0.8527 & 0.6029 & 0.4868 & -0.3901 & -0.6794 & 0.3389 & 0.6066 \\ 0.3746 & 1.0160 & -0.6758 & -0.0013 & -0.8689 & -0.0004 & -0.5348 & 0.3186 \end{bmatrix} \tag{13}$$

$$Input = \begin{bmatrix} CSC \\ TSC \\ CA \\ D_{max} \\ SPC \\ FM \\ W/B \\ SR \end{bmatrix} \quad (14)$$

$$b1 = \begin{bmatrix} -1.6649 \\ -0.5550 \\ 0.5550 \\ 1.6649 \end{bmatrix} \quad (15)$$

$$b2 = [-0.5543] \quad (16)$$

## 6 Conclusions

The present research aimed to propose a novel hybrid model for the early prediction of the uniaxial compressive strength of concrete. To this end, satin bowerbird optimizer was examined as the main algorithm, along with Henry gas solubility optimization, sunflower optimization, and vortex search algorithm as benchmarks. Four hybrid models were created by synthesizing these algorithms with an artificial neural network. Next, the models were evaluated in capturing and reproducing the UCS behavior. Overall, the findings demonstrated that the performance of the ANN-SBO and ANN-VSA with R > 0.95, is more promising than two other hybrids (i.e., ANN-HGSO and ANN-SFO) in both phases. Also, more detailed evaluations (MAPE of 8.0629 vs. 9.4970 %) showed the superiority of the ANN-SBO. All in all, this algorithm could strongly analyze the relationship between the UCS and influencing factors and tune the neural parameters accordingly. Thus, the SBO-based hybrid can be an efficient substitute for laboratory and traditional approaches.